\documentclass[letterpaper, 10 pt, conference]{ieeeconf}

\IEEEoverridecommandlockouts
\usepackage[T1]{fontenc}
\usepackage{aecompl}

\usepackage{cite,soul}
\usepackage{amsmath,amssymb,amsfonts}
\usepackage{algorithmic}
\usepackage{graphicx}
\usepackage{subfigure}
\usepackage{textcomp}
\usepackage{xcolor}

\usepackage[ruled]{algorithm2e}
\usepackage{booktabs}
\usepackage{multirow}
\usepackage{epstopdf}
 
\newtheorem{remark}{Remark}

\def\BibTeX{{\rm B\kern-.05em{\sc i\kern-.025em b}\kern-.08em
    T\kern-.1667em\lower.7ex\hbox{E}\kern-.125emX}}

\title{\LARGE \bf
Curriculum Proximal Policy Optimization with Stage-Decaying Clipping for Self-Driving at Unsignalized Intersections
}

\author{Zengqi Peng, Xiao Zhou, Yubin Wang, Lei Zheng,  Ming Liu, \textit{Senior Member, IEEE,} and Jun Ma
\thanks{This work was supported in part by the Guangzhou-HKUST(GZ) Joint Funding Scheme under Grant 2023A03J0148; and in part by the Project of Hetao Shenzhen-Hong Kong Science and Technology Innovation Cooperation Zone under Grant HZQB-KCZYB-2020083.}
\thanks{Zengqi Peng, Xiao Zhou, Yubin Wang, and Lei Zheng are with the Robotics and Autonomous Systems Thrust, The Hong Kong University of Science and Technology
(Guangzhou), Guangzhou, China. }
\thanks{Ming Liu and Jun Ma are with the Robotics and Autonomous Systems Thrust, The Hong Kong University of Science and Technology (Guangzhou), Guangzhou, China, also with the Department of Electronic and Computer Engineering, The Hong Kong University of Science and Technology, Hong Kong SAR, China, and also with the HKUST Shenzhen-Hong Kong Collaborative Innovation Research Institute, Futian, Shenzhen, China.}
\thanks{All correspondence should be sent to Jun Ma (e-mail: jun.ma@ust.hk).}
}

\begin{document}

\maketitle

\thispagestyle{empty}
\pagestyle{empty}

\begin{abstract}
Unsignalized intersections are typically considered as one of the most representative and challenging scenarios for self-driving vehicles.
To tackle autonomous driving problems in such scenarios, this paper proposes a curriculum proximal policy optimization (CPPO) framework with stage-decaying clipping. 
By adjusting the clipping parameter during different stages of training through proximal policy optimization (PPO), the vehicle can first rapidly search for an approximate optimal policy or its neighborhood with a large parameter, and then converges to the optimal policy with a small one. 
Particularly, the stage-based curriculum learning technology is incorporated into the proposed framework to improve the generalization performance and further accelerate the training process. 
Moreover, the reward function is specially designed in view of different curriculum settings. 
A series of comparative experiments are conducted in intersection-crossing scenarios with bi-lane carriageways to verify the effectiveness of the proposed CPPO method. 
The results show that the proposed approach demonstrates better adaptiveness to different dynamic and complex environments, as well as faster training speed over baseline methods.

\end{abstract}

\section{Introduction}

In the past few decades, both academia and industry have witnessed the rapid development of autonomous driving technology \cite{paden2016survey,narayanan2020shared,kiran2021deep}. However, ensuring safe and efficient passage at intersections with high vehicle density and frequent vehicle interactions remains a challenging task for autonomous driving \cite{wei2021autonomous}, particularly in the presence of numerous human-driven vehicles exhibiting unpredictable behaviors. Inaccurate prediction of surrounding vehicle behavior can certainly influence the decision-making of the autonomous vehicle and even pose a threat to its safety. The situation becomes even more complicated when it comes to the unsignalized intersection, where the autonomous vehicle could interact with surrounding vehicles from multiple different directions simultaneously. In this sense, the increasing number of surrounding vehicles and their mutual influences lead to more complex behavior modes that are challenging to forecast, thus severely affecting the safety and travel efficiency of the autonomous vehicle.

Currently, substantial research efforts focus on autonomous driving development, including rule-based methods, optimization-based methods, and learning-based methods. 
As a representative of the approach, rule-based methods show the promising effectiveness due to their transparency and comprehensibility. A set of rules are proposed to clarify the sequence of vehicles traversing the unsignalized intersection scenario. With planned rules, each vehicle makes the decision for preempting or yielding the surrounding vehicles to guarantee the road traffic safety \cite{lu2014rule}. Generally, such a rule-based strategy is designed to prioritize road traffic safety and avoid potential collisions with other social vehicles at any cost \cite{aksjonov2021rule}.
Moreover, optimization-based methods, such as model predictive control (MPC), are also widely utilized owing to their effectiveness in generating the control strategy while dealing with various constraints \cite{qian2016motion,wang2023chance}. 
An autonomous and safe intersection-crossing strategy is developed in \cite{riegger2016centralized} where the optimized trajectories of a team of autonomous vehicles approaching the intersection area are generated by centralized MPC. 
An effective intersection-crossing algorithm for autonomous vehicles based on vehicle-to-infrastructure communication capability is proposed in \cite{kneissl2018feasible}, where all vehicles are navigated by decentralized MPC with the sharing setting of the expected time of entering a critical zone. However, rule-based and optimization-based methods suffer from adaptiveness to the time-varying traffic situation due to the high complexity and dynamicity in real-world driving scenarios.

\begin{figure*}[!t] 
    \centering   
    \includegraphics[trim=0.5cm 0 0 0, width=0.8\linewidth]{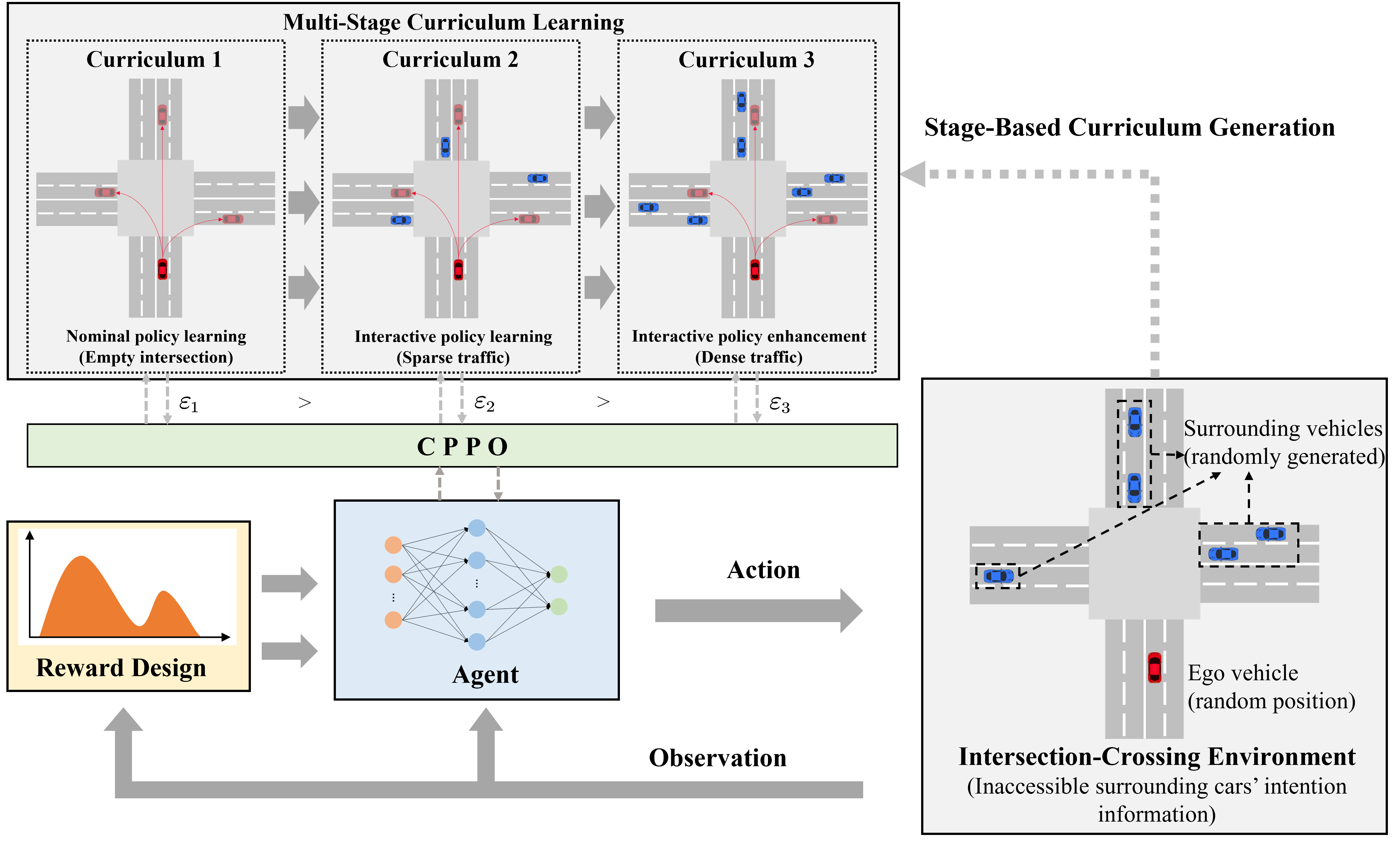}
   \caption{Overview of the proposed CPPO framework for autonomous driving at unsignalized intersection. In the four-way intersection scenario, the ego vehicle is depicted in red and the surrounding vehicles are in blue. The solid vehicle represents the start point, and the semi-transparent vehicle represents the goal point. 
   }
    \label{frame}
\end{figure*}

On the other hand, learning-based methods have recently been developed in the field of robotics and autonomous driving. Particularly, imitation learning-based methods
leverage expert experience data to train the agent to generate trajectory or control commands \cite{codevilla2018end,menda2019ensembledagger}. However, as supervised learning methods, the quality of the expert demonstration dataset will influence the actual performance of the agent significantly, which makes the training process rather challenging. 
Reinforcement learning (RL) is a promising direction to handle self-driving tasks, such as deep Q-learning, soft actor-critic, and proximal policy optimization (PPO) \cite{sallab2017deep,kiran2021deep,schulman2017proximal}. The target of RL-based methods is to train the agent to obtain the policy that maximizes the future cumulative reward by exploring the environment. A deep RL framework is proposed for navigation at occluded intersections by combining the deep Q-network and time-to-go representation \cite{isele2018navigating}, which demonstrates higher travel efficiency and lower collision rate than the time-to-collision method. A hierarchical decision algorithm is proposed for self-driving at intersections by integrating an RL-based decision module and an MPC-based planner \cite{tram2019learning}. However, this work only considers the situation of vehicles traveling straight, yet considering the situation where the vehicle turns left or right will significantly increase the complexity of the problem. In \cite{zhou2019development}, an RL-based car following model is proposed for connected and automated vehicles at signalized intersections. The arrival time prediction is introduced into the reward function to train the agent.

Nevertheless, a major drawback of the RL-based method is that it always requires a long training process to obtain acceptable driving policies for complex self-driving tasks, such as unsignalized intersection-crossing. Because the training environment is unknown to the agent, which leads to that the agent needs to spend plenty of time interacting with the environment to understand its characteristics before learning a satisfying strategy. To solve this problem, a model-accelerated PPO algorithm is proposed in  \cite{guan2020centralized}, where a prior model is incorporated into the RL framework to accelerate the training process. However, due to the black-box nature of neural networks, the safety of policies cannot be strictly guaranteed. Transfer learning is a class of techniques to leverage external expert knowledge before the learning process of target tasks, which helps to speed up the training procedure \cite{zhu2020transfer}. Moreover, curriculum learning is an alternative solution to expedite network training, which initiates the training process from easier tasks \cite{wang2021survey,narvekar2020curriculum}. Specifically, in curriculum learning, a series of course tasks with increasing difficulties are designed to enable agents to learn optimal strategies faster and more efficiently. In \cite{song2021autonomous}, the curriculum learning technology is introduced into the soft actor-critic algorithm for autonomous overtaking tasks, which leads to faster convergence speed compared to the vanilla RL method. Besides, an automated curriculum mechanism is proposed in \cite{khaitan2022state} to train agents for traversing the unsignalized intersection. The agent can obtain a fine-tuned policy in the final phase by dropping the future state information during the training process. However, in this work, surrounding vehicles are assumed not to interact with other vehicles, and future trajectories of opposing vehicles are accessible. 
Besides, the total number of surrounding vehicles is fixed. These assumptions and settings could possibly limit the generalization of the trained policy in more dynamic intersection scenarios.

This work addresses the unsignalized intersection-crossing task where the ego vehicle interacts with varying numbers of surrounding vehicles. The main contributions of this work are summarized as follows: 
a curriculum proximal policy optimization (CPPO) framework with stage-decaying clipping is proposed for training the agent in the highly dynamic intersection-crossing self-driving task, where the reward function is particularly designed to balance the safety and travel efficiency in different traffic situations. The stage-based curriculum learning technology is introduced into the PPO method with a decaying clipping parameter to accelerate the training process and improve the generalization of the trained policy. By learning a series of courses with increasing difficulty levels, the agent can capture the uncertainties of surrounding vehicles implicitly and adapt to situations effectively with varying numbers of surrounding vehicles. A series of simulations in different intersection scenarios are conducted to evaluate the performance of the proposed method and baseline method in $Highway\_Env$ \cite{highway-env}. The CPPO method demonstrates faster training speed and better generalization than the standard PPO method. 

The rest of this paper is organized as follows. Section II gives an introduction to the formulation of the intersection problem addressed in this work. Section III illustrates the proposed methodology. Section IV presents the experimental results. The conclusion and future work are discussed in Section V.

\section{Problem Definition}
In this section, we first introduce the task scenarios to be solved in this work. Then, the formulation of the 
learning environment is illustrated.
\subsection{Problem Statement}
The problem to be solved in this work is to control the ego vehicle to traverse an unsignalized four-way intersection and reach the goal position. Furthermore, each road consists of two lanes. The task scenario is shown in Fig.~\ref{frame}.

We assume that the ego vehicle always starts from a random position (denoted by a solid red vehicle) in the lower zone of the intersection, and the goal position of the ego vehicle (denoted by a semi-transparent red vehicle) is also randomly generated within the left, upper, and right zones. There are several surrounding vehicles driving from other lanes towards different target lanes. 
They will react to the behavior of the ego vehicle. 
Without loss of generality, we assume that the position and velocity information of surrounding vehicles can be accessed by the ego vehicle. But the information about the driving intention of surrounding vehicles is unknown to the ego vehicle, which increases the difficulties of this task. 
The objective of this task is to generate a sequence of actions that enables the ego vehicle to expeditiously approach the target point while ensuring collision avoidance with surrounding vehicles and staying within the road boundaries.

\subsection{Learning Environment}

In this work, we frame the agent's learning objective as the optimal control of a Markov Decision Process by defining state space, action space, state transition dynamics, reward function, and discount factor. Then the RL problem can be represented by a tuple $\mathcal{E} = \langle \mathcal{S}, \mathcal{A}, \mathcal{P}, \mathcal{R}, \gamma \rangle$.

\textbf{State space $\mathcal{S}$}: In this scenario, the state space $\mathcal{S}$ is explicitly defined by $Highway\_Env$ environment, which consists of state matrices $\mathbf{S}_t$. 
At each timestep $t$, the agent observes the kinematic features of vehicles in the intersection, so the state matrix can be defined as follows,
\begin{equation}
\begin{split}
\mathbf{S}_t=\left[\ \mathbf{s}_t^0\ \ \mathbf{s}_t^1\ ...\ \mathbf{s}_t^N\ \right]^T
\end{split}
\label{state martix}
\end{equation}
where $N$ denotes the total number of vehicles in the intersection. The first row of $\mathbf{S}_t$, i.e., $(\mathbf{s}_t^0)^T$, is a vector consisting of the kinematic features of the ego vehicle, while other rows of $\mathbf{S}_t$, i.e., $(\mathbf{s}_t^i)^T\ (i=1,2,...,N)$, represent vectors of the kinematic features of the surrounding vehicles. The vector of the kinematic features is defined as follows,
\begin{equation}
\begin{split}
\mathbf{s}_t^{i}={\left[\begin{array}{l l l l l l l}{x_t^{i}}&{y_t^{i}}&{v_{x,t}^{i}}&{v_{y,t}^{i}}&{\sin\psi_t^{i}}&{\cos\psi_t^{i}}\end{array}\right]}^{T}
\end{split}
\label{kinematic_features}
\end{equation}
where $x_t^{i}$ and $y_t^{i}$ are the vehicle’s current position in the world coordinate system, respectively; $v_{x,t}^{i}$ and $v_{y,t}^{i}$ are the speed of the vehicle along the X-axis and Y-axis, respectively; $\psi_t^{i}$ is the heading angle of the vehicle at timestep $t$.

\textbf{Action space $\mathcal{A}$}: Similar to the human driver’s operation in driving, the action space of the agent consists of five basic discrete actions as follows,
\begin{equation}
\begin{split}
\mathcal{A}=\left\{ A^{0},A^{1},A^{2},A^{3},A^{4} \right\}
\end{split}
\label{Action space}
\end{equation}
where $A^{0}$ and $A^{2}$ are the left and right lane-changing action, respectively; $A^{1}$ is the motion keeping action; $A^{3}$ and $A^{4}$ represent the deceleration and acceleration action, respectively.

The vehicle is controlled by two low-level controllers, the longitudinal and lateral controllers, which convert discrete actions (\ref{Action space}) to the continuous control input $ [ a,\delta ]^{T}$ of the vehicle, where $a$ and $\delta$ are the acceleration and steering angle, respectively. Considering the physical limitations, the maximum acceleration and steering angle are set as 8 m/s$^{2}$ and $45$ degrees, respectively.

\textbf{State transition dynamics $\mathcal{P}(\mathbf{S}_{t+1}|\mathbf{S}_t,a_t)$}: The transition function $\mathcal{P}$ defines the transition of environment state, which follows the Markov transition distribution. The next state generated by $\mathcal{P}$ is related to the current state $\mathbf{S}_t$ and the applied action $a_t \in \mathcal{A}$. The transition dynamics $\mathcal{P}(\mathbf{S}_{t+1}|\mathbf{S}_t,a_t)$ is implicitly defined by $Highway\_Env$ environment and unknown for the agent.

\textbf{Reward function $\mathcal{R}$}: The reward function assigns a positive reward for successfully completing an episode and for maintaining survival. 
It penalizes collisions, out-of-the-road, and lane-changing behaviors. When there are few vehicles on the road, the acceleration behaviors will be rewarded, and vice versa. The details about the structure of the reward function will be introduced in Section III-C.

\textbf{Discount factor $\gamma \in (0,1)$}: The future reward is accumulated with a discount factor $\gamma$.

\section{Methodology}

In this section, we first illustrate the details of the proposed CPPO framework with stage-decaying clipping. Then, the curriculum setting is introduced to enhance the training process of the RL agent. Lastly, the multi-objective reward structure for the agent is presented.

\subsection{Curriculum Proximal Policy Optimization with Stage-Decaying Clipping}

PPO is a model-free RL framework to solve the sequential decision-making problem under uncertainties. It alternatively constructs a clipped surrogate objective to replace the original function, which generates a lower bound on the unclipped objective and avoids the incentive for an excessively large policy update. Therefore, PPO algorithm facilitates the learning of
policies in a faster and more efficient way. 
Specifically, the objective function in the PPO algorithm is shown as follows,

\begin{equation}
J_{clip}(\theta)=\mathbb{E}_t\left[\min \left(\rho_t(\theta) \hat{A}_t, \operatorname{clip}\left(\rho_t(\theta), 1-\varepsilon, 1+\varepsilon\right) \hat{A}_t\right)\right]
\end{equation}
where $\rho_t(\theta)=\frac{\pi_\theta(a_t|s_t)}{\pi_{\theta_{old}}(a_t|s_t)}$ is the probability ratio between the new policy and old policy, $\hat{A}_t$ is the estimated advantage function at timestep $t$, $\varepsilon$ is a hyperparameter.

The intuitive idea is to change the value of the hyperparameter $\varepsilon$ in the different training periods. Empirically, we set $\mathbf{\varepsilon}=\left\{0.25,0.2,0.15\right\}$. During the beginning stage of training, a large parameter $\varepsilon_1=0.25$ is used for the rough exploration, which is then adjusted to the second largest parameter $\varepsilon_2=0.2$ during the middle stage, and finally adjusted to a smaller parameter $\varepsilon_3=0.15$ during the later stage. However, determining when to make adjustments to the parameters is a problem.

\begin{remark}
If $\hat{A} >0$, the probability ratio will not exceed $1+\varepsilon$. Otherwise, the ratio will be less than or equal to $1-\varepsilon$. Therefore, the magnitude of the clipping parameter $\varepsilon$ will influence the training speed and the performance of the trained policy. 
A larger clipping parameter allows for a more significant step size of the update, which means that the agent can have a faster training process, but the optimality of the trained policy cannot be ensured.
On the contrary, a smaller clipping parameter may lead to slow update speed and drop into a local optimal policy. Therefore, stage-decaying clipping can capitalize on the strengths of both approaches while circumventing their weaknesses.
\end{remark}

In intersection-crossing tasks, the ego vehicle needs to interact with a variable number of interactive surrounding vehicles with different driving behaviors from the other three directions. Therefore, these scenarios are rather complex. It is difficult to get a satisfactory driving policy by directly deploying the PPO algorithm in these high-dynamic scenarios. Here, we introduce stage-based curriculum learning technology to generate a task sequence with increasing complexity for training acceleration and better generalization. Besides, the clipping parameter can be adjusted when switching the curriculum, which addresses the issue of when to change the clipping parameter mentioned before.

We generate a curriculum sequence with three stages, which is represented as $\mathbf{\Omega}=\left\{\Omega_1,\Omega_2,\Omega_3\right\}$. The curriculum sequence is designed for different goals with increasing complexity.

\textit{Curriculum 1}: \emph{Intersection without surrounding vehicles.} In stage 1, which is denoted as $\Omega_1$, there is only the ego vehicle in the intersection. The objective of this curriculum is to learn a transferable nominal policy to obtain the nominal policy $\pi_1$ that can find an action sequence to the goal point. The ego agent is guided with empirically designed rewards in this curriculum stage, which can decrease the exploration time in the whole action space and avoid local optimum with poor generalization. In this curriculum, the hyperparameter of the clipped function is set as $\varepsilon_1$.

\textit{Curriculum 2}: \emph{Intersection with a few vehicles.} In the second stage, we load the policy trained in Curriculum 1 as the initial policy of this stage for the following training process. In this curriculum, there are a few surrounding vehicles in the intersection. We aim to train the nominal policy to obtain a new policy $\pi_2$ with the preliminary obstacle avoidance ability by maximizing the intersection-crossing reward in this stage. The hyperparameter is switched to $\varepsilon_2$.

\textit{Curriculum 3}: \emph{Intersection with numerous vehicles.} In the third stage, we load the policy trained in Curriculum 2 as the initial policy. In this stage, there are numerous surrounding vehicles in the intersection. The objective of this curriculum is to train the previous policy to obtain the optimal policy $\pi^*$ with better obstacle avoidance ability in a more complex environment. In the preceding episodes of this stage, the parameter $\varepsilon$ remains at 0.2, and then transits to $\varepsilon_3$ to continue the course training.

\subsection{Multi-Objective Reward Design}
For the RL-based method, the design of the reward is essential for the success of policy training. An inappropriate reward function not only slows down the speed of training but also leads to a trained policy with poor performance. Therefore, it is challenging to guide the agent to obtain a satisfactory driving strategy in a complex scenario.
In this work, the reward is smartly designed for intersection scenarios with different densities.

Considering the complexity of the target scenario, a comprehensive reward function is designed as follows,

\begin{equation}
\begin{split}
    r &=  r_{succ}(T,N_{car}) + r_{colli}(v, N_{car})\\ 
    & \quad + r_{TO} + r_{OfR} + r_{LC} + r_{l},
\end{split}
\label{reward_func}
\end{equation}
where $r_{succ}$ and $r_{l}$ are the reward for successfully completing tasks and surviving in the task, respectively; $r_{colli}, r_{TO}, r_{OfR},$ and $r_{LC}$ are the penalty of collision with surrounding vehicles, time-out, out-of-the-road boundary, and lane-changing behavior, respectively.

\begin{remark}
Inspired by the idea of curriculum learning, several terms in the reward function are related to the setting of the scenario. The success reward term is related to both the time of finishing the task and the maximum number of surrounding vehicles in the intersection central area when the ego vehicle is crossing. It will encourage autonomous vehicles to expedite the intersection-crossing task, and greater rewards will be obtained when completing more complex tasks. Additionally, the collision penalty is related to the speed of the ego vehicle when the collision happens and the maximum number of surrounding vehicles. This reward term aims to encourage autonomous vehicles to maintain a lower speed when there are more vehicles in the central area of the intersection to ensure safety, and greater penalties will be imposed if collisions occur in more complex tasks. Thereby, the balance between the safety and travel efficiency is achieved through the particular designed reward function.
\end{remark}

To sum up, the proposed CPPO framework with stage-decaying clipping is summarized in Algorithm 1.

\begin{algorithm}[t]  
	\caption{Curriculum Proximal Policy Optimization with Stage-Decaying Clipping} \label{ALG_CPPO}
	\LinesNumbered 
	\KwIn{Environment state $s_t$, curriculum set $\mathbf{\Omega}$}
	\KwOut{$\pi^* = f(\mathbf{\theta}^*)$}
        Initialize the policy network with parameter $\mathbf{\theta_0}$\; 
        \While{not terminated}  
            {Select curriculum $\Omega_i$ from curriculum set $\mathbf{\Omega}$\;
		  Reset the environment according to the setting of curriculum $\Omega_i$\;
            Select the clipping parameter $\varepsilon = \varepsilon_j$\;
            \If{curriculum switched}
                {Load policy $\pi^*$ trained by $\Omega_{i-1}$ as the initial policy\;}
                Update the policy network by maximizing the designed reward (\ref{reward_func})\;
                Save the trained policy as $\pi_i$;}
            The policy network obtained by the last curriculum is the final policy $\mathbf{\theta}^* = \mathbf{\theta_3}$.\
\end{algorithm}

\section{Experiments}
In this section, we implement the proposed algorithm in dynamic intersection-crossing scenarios. Then we compare the performance of CPPO with that of two baseline methods. The simulations are conducted in $Highway\_Env$ \cite{highway-env}.

\subsection{Experimental Settings}

The experiments are conducted on the Windows 11 system environment with a 3.90 GHz AMD Ryzen 5 5600G CPU. The task scenarios are constructed based on $Highway\_Env$ environment, where each road is a bi-lane carriageway. We use fully-connected networks with 1 hidden layer of 128 units (action network) and 64 units (critic network) to represent policies. The neural network is constructed in PyTorch \cite{paszke2019pytorch} and trained with an Adam optimizer \cite{kingma2014adam}.
The MDP is solved using the proposed CPPO framework and the standard PPO method with two different fixed clipping parameters. The simulation frequency is set as $15$ Hz. The hyperparameters for network training are listed in Table ~\ref{simu_setup}. Here, we compare the proposed method, CPPO, and two baseline methods with different clipping parameters ($\varepsilon=0.15, 0.25$). For the sake of fairness, other network parameters of these methods are the same.

\begin{table}[htbp]
    \centering
    \caption{Hyperparameter settings.}
    \label{simu_setup}
    \begin{tabular}{cc}
        \toprule
        Hyperparameter  & Value \\
        \midrule
        Learning rate for actor network  & $5 \times 10^{-4}$ \\
        Learning rate for critic network  & $1 \times 10^{-3}$ \\
        Discount factor  & 0.9 \\
        Number of epochs  & 20 \\
        \bottomrule
    \end{tabular}
\end{table}

Then we test all trained policies in intersection scenarios with different numbers $N_{sv}$ of surrounding vehicles, whose behaviors are characterized by the intelligent driver model (IDM) \cite{treiber2000congested}. We record the success rate, collision rate, time-out rate, and out-of-road rate, respectively. By conducting these simulations, we can evaluate the generalization performance of these trained policies.
\subsubsection{No surrounding vehicles ($N_{sv}=0$)} This scenario can be used to check whether the trained policy has the ability to find the nominal trajectory to achieve the goal point.

\subsubsection{Different number of surrounding vehicles ($N_{sv}=1,2,...,6$)} These trained policies are tested in simple and complex scenarios to estimate their generalization performance and safety.

\begin{figure}[!htbp]
\centerline{\includegraphics[trim={0 0 0 0.8cm},width=0.45\textwidth]{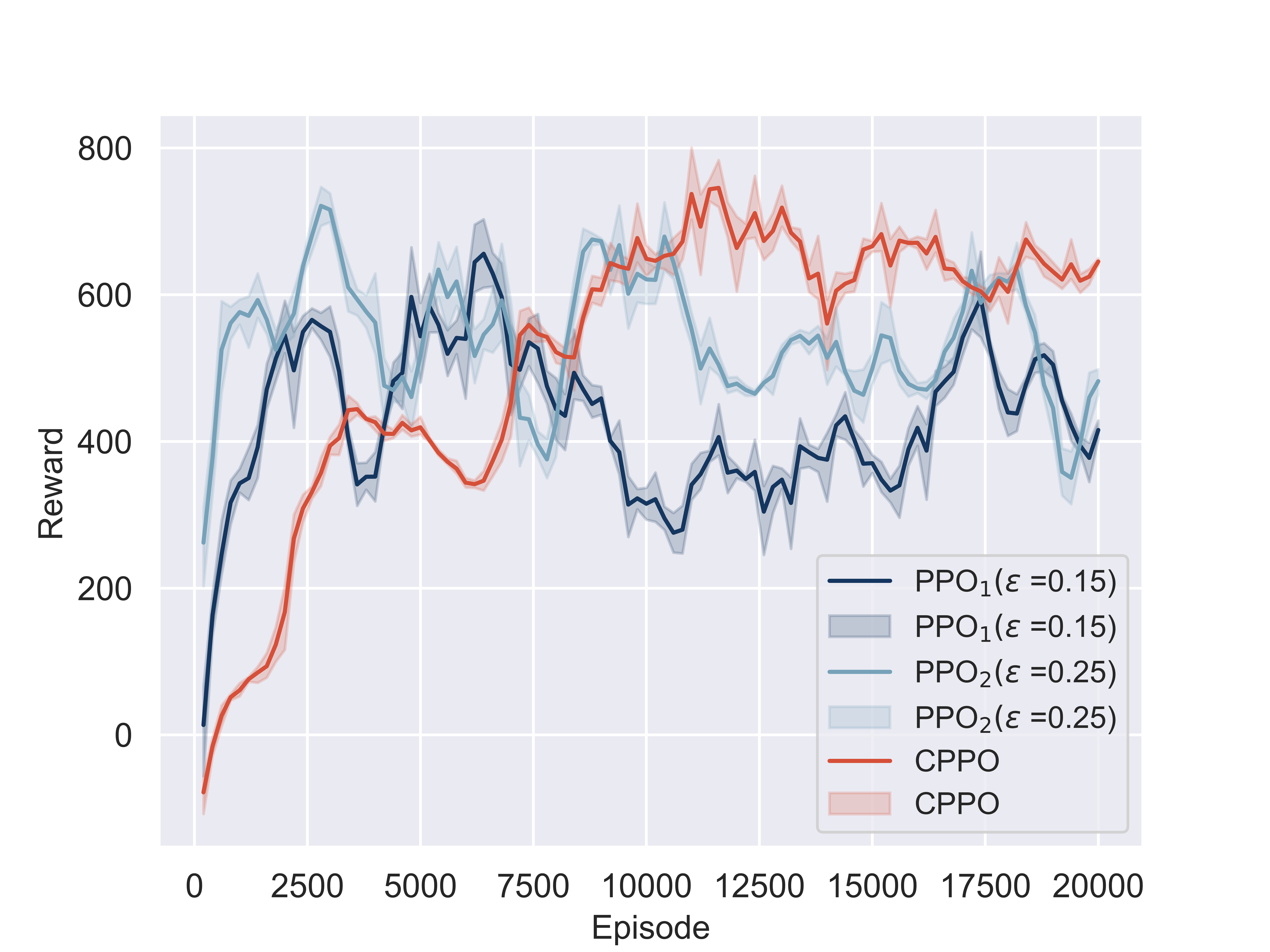}} 
\caption{Reward curve comparison among different methods. The training curves are smoothed by the Savitzky-Golay filter. The curriculum is switched at $2\times 10^3$th and $5\times 10^3$th episode, respectively.}
\label{reward}
\end{figure}

\begin{table}[htbp]
    \centering
    \caption{Training Time of Different Methods.}
    \label{train_time}
    \begin{tabular}{cc}
        \toprule
        Methods  & Training Time (hour:min.:sec.) \\
        \midrule
        CPPO  & 3:34:28  \\
        PPO$_1$ ($\varepsilon = 0.15$)  & 6:45:32 \\
        PPO$_2$ ($\varepsilon = 0.25$)  & 5:58:28 \\
        \bottomrule
    \end{tabular}
\end{table}

\begin{figure*}[htbp]
\centering
\subfigure[Timestep $t=5$]{
\begin{minipage}[t]{0.24\linewidth}
\centering
\includegraphics[width=0.9\linewidth]{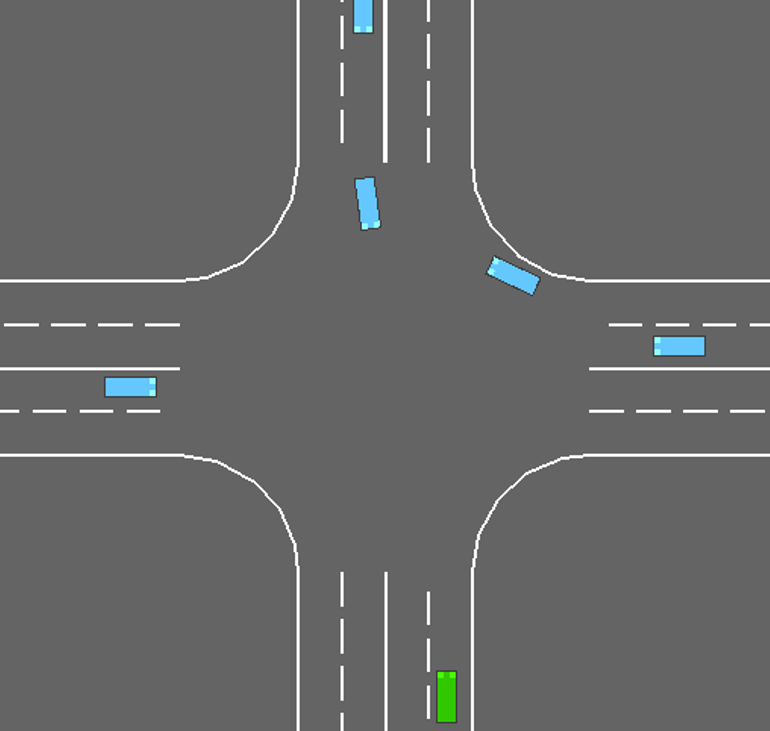}
\end{minipage}%
}%
\subfigure[Timestep $t=25$]{
\begin{minipage}[t]{0.24\linewidth}
\centering
\includegraphics[width=0.9\linewidth]{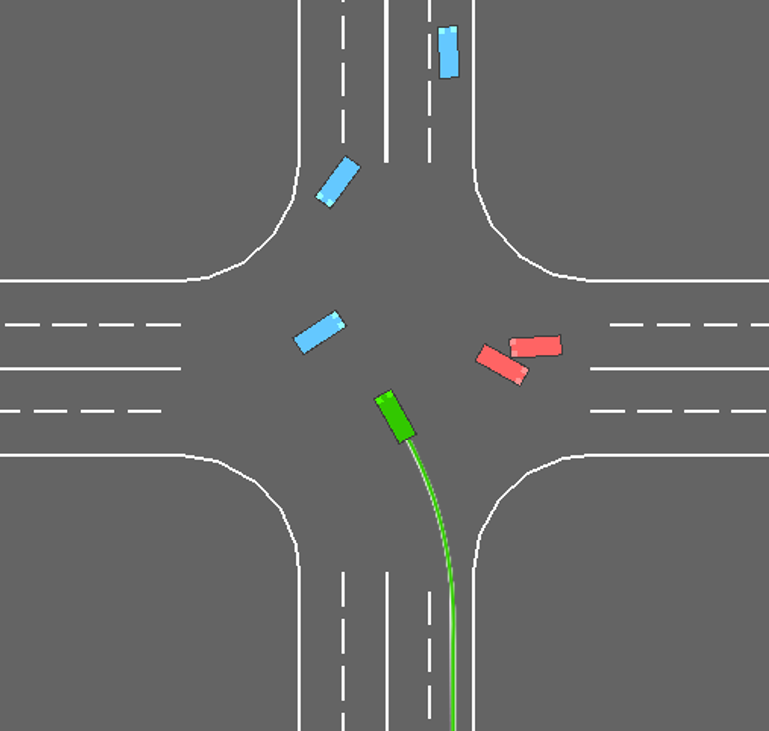}
\end{minipage}%
}%
\centering
\subfigure[Timestep $t=35$]{
\begin{minipage}[t]{0.24\linewidth}
\centering
\includegraphics[width=0.9\linewidth]{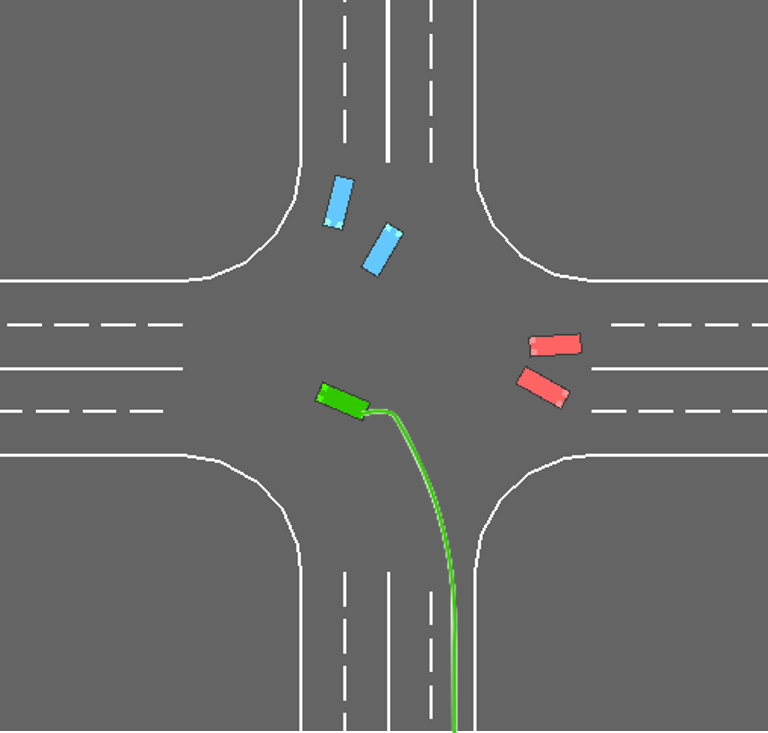}
\end{minipage}%
}
\subfigure[Timestep $t=55$]{
\begin{minipage}[t]{0.24\linewidth}
\centering
\includegraphics[width=0.9\linewidth]{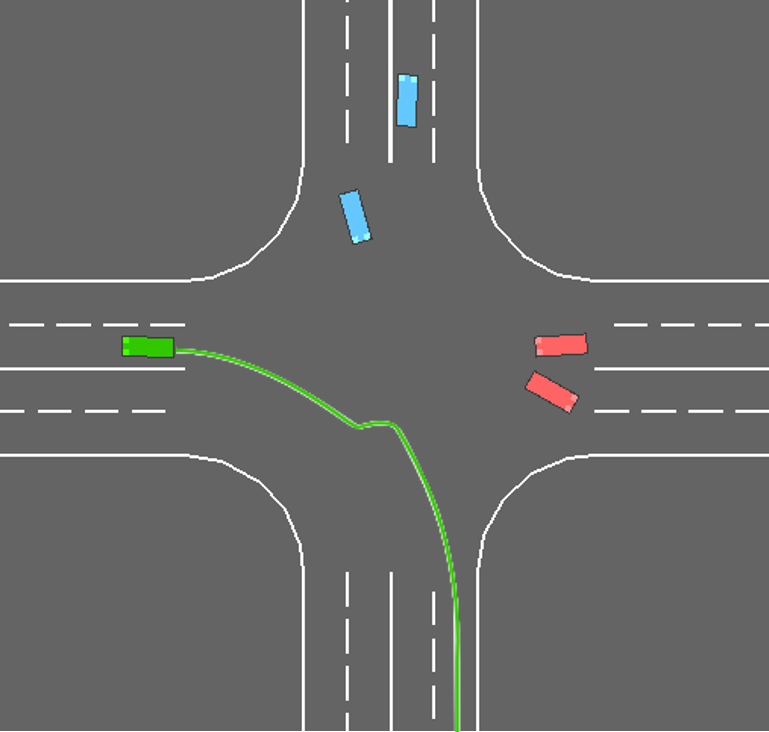}
\end{minipage}%
}
\caption{Demonstration of the driving performance attained by the proposed CPPO method in an unprotected left-turn task at the unsignalized intersection. The green car and blue cars represent the ego vehicle and surrounding vehicles under normal driving conditions, respectively. The red cars represent the vehicles that have collided. (a)-(d) present four key snapshots during the performance evaluation demonstration.}
\label{demonstration_traj}
\end{figure*}

\begin{table*}[]
\centering
\caption{Comparison of success rate, collision rate, and time-out rate among different methods.
}
\label{test_result}
\begin{tabular}{c|ccc|ccc|ccc}
\hline
\multirow{2}{*}{Methods} & \multicolumn{3}{c|}{$N_{sv}$=0}            & \multicolumn{3}{c|}{$N_{sv}$=2}             & \multicolumn{3}{c}{$N_{sv}$=4}    \\ \cline{2-10} 
                         & succ.(\%) & coll.(\%) & time-out(\%) & succ.(\%) & coll.(\%) & time-out(\%) & succ.(\%) & coll.(\%) & time-out(\%)  \\ \hline
CPPO                     & 100         & 0         & 0            & 90.5         & 9.5         & 0            & 78.5         & 21.5         & 0            \\
PPO$_1$($\varepsilon = 0.15$)                     & 55.5         & 0         & 44.5            & 86        & 14         & 0            & 76.5         & 23.5         & 0            \\
PPO$_2$($\varepsilon = 0.25$)                     & 69.5         & 0         & 30.5            & 83.5          & 16.5         & 0            & 72         & 28         & 0            \\ \hline
\end{tabular}
\end{table*}

\subsection{Training Results}
To illustrate the effectiveness of the proposed framework, the training time is listed in Table \ref{train_time} for comparison. It is obvious that CPPO has a much faster training speed than the two baseline methods. Specifically, the training speed of CPPO is $47.2\%$ faster than PPO$_1$, and $40.2\%$ faster than PPO$_2$. 

The change in reward during the training process is shown in Fig. \ref{reward}. According to these three learning curves, we can find that the CPPO agent initially receives the least reward, which is because that the reward function is positively correlated with the number of vehicles in the environment upon successful task completion. However, as CPPO's policy network converges to the optimal policy in the final stage, its reward curve surpasses that of the other two baseline methods. 
As the baseline algorithm PPO$_1$, with the parameter $\varepsilon=0.15$, is deployed directly in a complex environment for learning, its reward curve exhibits minor fluctuations in the later stages of training and is lower than that of the CPPO algorithm. For the agent trained by the baseline PPO$_2$ with $\varepsilon=0.25$, its reward curve exhibits a rapid increase initially. However, due to the large parameter $\varepsilon$, it oscillates significantly in later episodes. Although its reward curve ends up resembling that of PPO$_1$, the performance of its policy may be inferior to that of PPO$_1$. 
Above results illustrate that the introduction of stage learning allows for a more efficient sampling process, leading to a faster training speed and better convergence compared to baseline methods.

\subsection{Performance Evaluation}
To further demonstrate the superiority of CPPO in unsignalized intersection scenarios, 
comparative simulations are conducted. We test policies obtained by three methods in intersection scenarios with different numbers of surrounding vehicles. Each method is retested 200 times in each scenario. Among all testing results attained by the proposed CPPO method at the unsignalized intersection, we pick up the results from a left-turn task for demonstration, and the details are presented in Fig.~\ref{demonstration_traj}. 
In this demonstration, the ego vehicle is generated on the right lane of the lower zone, and its goal lane is the left lane in the left zone. We can observe that the ego vehicle exhibits a safe interaction behavior of decelerating and steering left when encountering a surrounding vehicle approaching from its left side in Fig.~\ref{demonstration_traj}. Specifically, in the first snapshot, the ego vehicle drives at a constant speed from the right lane in the lower zone towards the central area of the intersection, preparing for a left turn. Then, in the second snapshot, the ego vehicle perceives that it is getting close to a surrounding vehicle ahead, and the vehicle ahead shows no signs of slowing down to yield. As a result, the ego vehicle decelerates and steers left to avoid a collision. Afterwards, in the third snapshot, the closest surrounding vehicle in the previous snapshot has moved away from the ego vehicle. The ego vehicle perceives that there are no other vehicles that could potentially collide with itself. Therefore, it adjusts its heading angle, accelerates towards the target lane, and continues driving until the completion of the unsignalized intersection task.

\begin{figure}[!htbp]
\centering{\includegraphics[trim={0 0 0 1cm},width=0.47\textwidth]{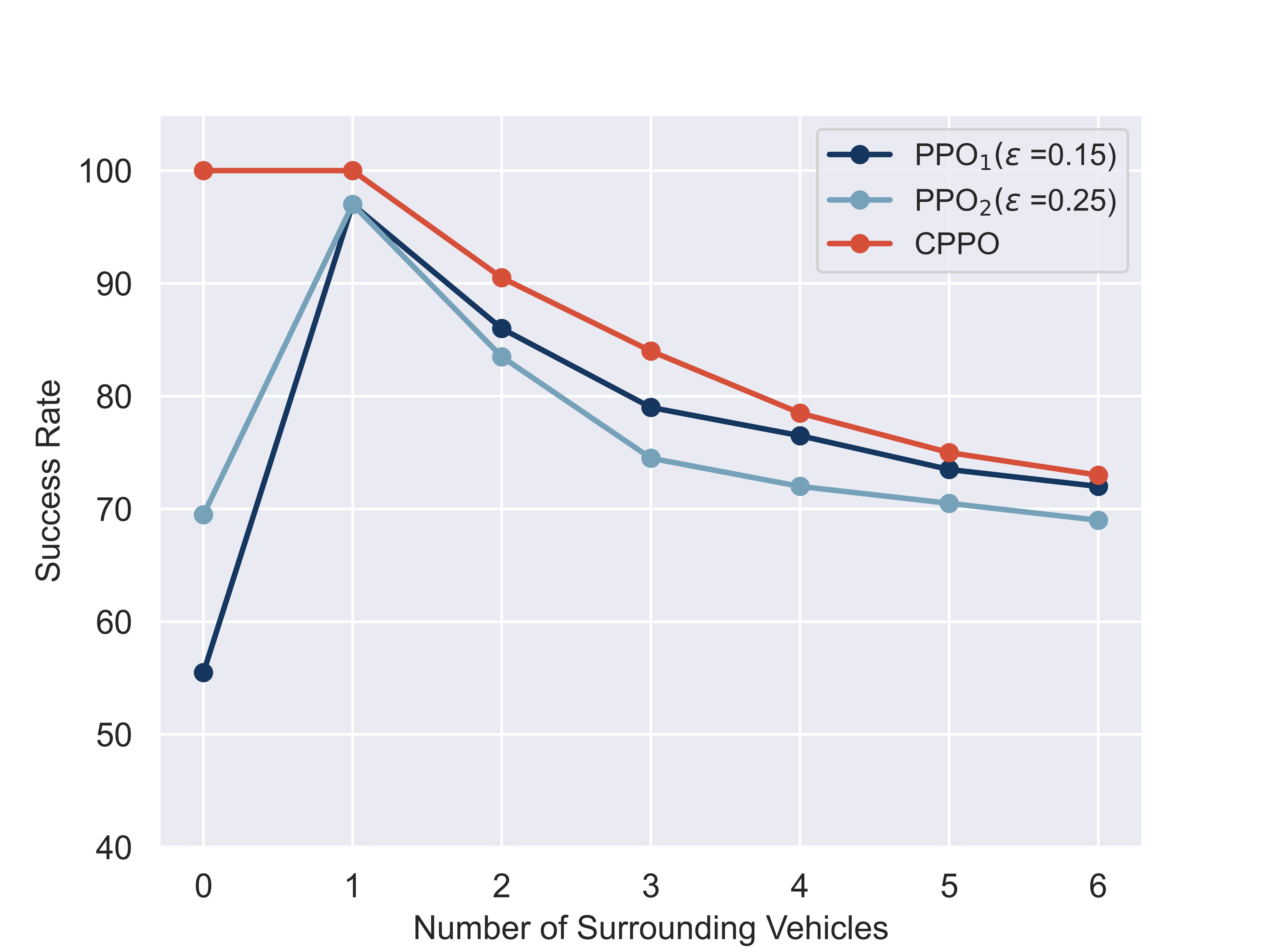}} 
\caption{Success rate comparison of three algorithms in unsignalized intersection with a different number of surrounding vehicles.}
\label{success_rate}
\end{figure}

The success rates of the three methods in all evaluation scenarios are shown in Fig. \ref{success_rate}. It is evident that the CPPO method achieves the highest overall task success rate. While its success rate decreases with the increasing complexity of the environment, it remains higher than that of the other two baseline methods. It indicates that the proposed method has better generalization performance. Furthermore, except for the scenario where there are no surrounding vehicles, PPO$_1$ achieves a slightly higher task success rate than PPO$_2$. This is because a smaller parameter $\varepsilon$ enables a search for better policies. Besides, it is noted that both two baseline methods have a low task success rate in the scenario where there is only the ego vehicle.

In addition, we have summarized the results of the success rate, collision rate, and timeout rate of tests with $N_{sv}=0, 2, 4$ surrounding vehicles in Table \ref{test_result}. From this statistical result, we can observe that both baseline methods exhibit a large number of timeouts in testing, $44.5\%$ and $30.5\%$ for PPO$_1$ and PPO$_2$, 
where there is no surrounding vehicles. This suggests that directly deploying the agent in a complex environment for training may cause the policy to become stuck in a local optimum. For other task scenarios with surrounding vehicles, two baseline methods did not exhibit timeouts in testing results. Because of the integration of the curriculum sequence, there is no timeout case happening for the CPPO. Therefore, the introduction of the curriculum learning technologies enables the agent to converge to a better optimum compared to those agents trained directly.

\section{Conclusion}

In this paper, we proposed a novel CPPO framework with stage-decaying clipping for unsignalized intersection-crossing tasks. We formulate a curriculum sequence for guiding the agent to learn the driving policy in scenarios where task difficulty gradually heightens by increasing the number of surrounding vehicles, and the clipping parameter in PPO varies as the curriculum stage switches. Besides, the reward function is particularly designed to guide the agent to balance safety and travel efficiency in different situations. A series of experiments were conducted in $Highway\_Env$ environment to verify the effectiveness of the proposed method. We compared the performance of the proposed method and two baseline methods. The results show that the CPPO method has the fastest training speed and the highest task success rate among different settings, which demonstrates that the proposed method has better generalization performance than all baseline algorithms. In the future, we will consider incorporating game theoretic methods into the CPPO framework to enhance the effectiveness of our method.

\bibliographystyle{ieeetr}
\bibliography{reference}
\end{document}